\documentclass[11pt]{article}

\usepackage[preprint]{acl}

\usepackage{times}
\usepackage{latexsym}
\usepackage{amsmath}

\usepackage[T1]{fontenc}

\usepackage[utf8]{inputenc}

\usepackage{microtype}

\usepackage{inconsolata}

\usepackage{graphicx}
\usepackage{subcaption}

\usepackage{booktabs}

\usepackage{hyperref}

\usepackage{tikz,tikz-qtree,tikz-dependency}
\usetikzlibrary{snakes,arrows,shapes,backgrounds,calc}
\tikzstyle{internal}=[circle,draw=black,thick,inner sep=1.5pt]
\tikzstyle{internalnum}=[circle,draw=black,inner sep=0.75pt,font={\tiny}]
\tikzstyle{external}=[circle,draw,fill=black,inner sep=1.5pt]
\tikzstyle{externalnum}=[circle,draw,fill=black,inner sep=0.75pt,font={\color{white}\tiny}]
\tikzset{level distance=24pt,sibling distance=12pt}
\tikzset{edge from parent/.style={draw,-latex,edge from parent path={(\tikzparentnode)--(\tikzchildnode)}}}
\tikzset{llabel/.style={auto=right,inner ysep=1pt,inner xsep=2pt}}
\tikzset{rlabel/.style={auto=left,inner ysep=1pt,inner xsep=2pt}}
\tikzset{every leaf node/.style={anchor=north}}
\title{Does Topic Sentiment Cause Perceived Ideology? Comparing Human and LLM Annotations in Political News Articles}

\author{Upasana Chatterjee \\
Department of Computer Science \\
  Columbia University \\
  \texttt{uc2143@columbia.edu} \\
  }

\begin{document}
\maketitle
\begin{abstract}
We ask whether topic sentiment has a causal effect on perceived political ideology, and whether the answer depends on who assigns the ideology label. Using articles from AllSides, paired with shared sentiment annotations from Llama-3.3-70b-versatile, we compare ideology labels from expert human annotators, GPT-4o-mini (baseline and finetuned), and Llama-3.3-70B. We apply Double Machine Learning (DML) and mediation analysis across all four annotation paradigms. Zero-shot LLMs regularly inflate effect sizes relative to human annotations, while fine-tuning often attenuates them back toward the human scale. Our results have implications for the use of LLM annotations as silver labels and as proxies for human judgment in downstream causal analyses: they may be reliable for recovering the presence and direction of effects on the partisan topics, but not their magnitude, leading to over- or under-prediction of some ideology given particular topics.
\end{abstract}

\section{Introduction}
There is growing interest in using large language models (LLMs) as low-cost, scalable surrogates for human participants in social science research \citep{horton2023large, li2024frontiers}. This work suggests that LLMs may replicate aspects of human behavioral and judgmental responses in ways that make them useful proxies for study participants.

However, validating that LLM \textit{outputs} correlate with human outputs does not establish that LLMs arrive at those outputs through the same underlying processes. \citet{gao2025LLMscaution} provide a cautionary empirical demonstration: even in a simple economic game, advanced LLMs fail to replicate human behavior distributions, with failure modes that are inconsistent and difficult to predict. This raises a deeper question about \textit{causal fidelity}: whether the features that causally drive LLM predictions are the same features that causally drive human judgments. Existing validation studies assess output-level agreement; none, to our knowledge, test whether the causal structure of LLM decisions mirrors that of human decisions.

Concretely, focusing on topic-level sentiment as a specific, interpretable feature class, we define a causal question: \textit{What is the causal effect of topic sentiment on perceived ideological stance in news articles?} Following \citet{Pearl2001DirectAI}, we operationalize this question by defining appropriate counterfactuals and using a combination of NLP techniques and causal inference methods. We hold sentiment annotations constant across all experimental conditions and vary only the ideology labels, comparing expert human annotations from AllSides with predictions from GPT-4o-mini (baseline and finetuned) and Llama-3.3-70B. This design isolates differences attributable to the annotation source itself, enabling a direct test of causal fidelity.

We apply our framework to the AllSides expert-labelled articles ($N=9{,}830$) using Double Machine Learning \citep[DML;][]{chernozhukov2018} for causal effect estimation and mediation analysis \citep{Pearl2001DirectAI} to decompose direct and indirect pathways. We find that topic sentiment has a significant causal effect on predicted ideology on topics that are deemed explicitly partisan by the annotators. For example, sentiment toward Republican-coded topics shifts predictions rightward, and toward Democratic-coded topics leftward. Even for topics with significant effects across all paradigms, the magnitude of the causal effect varies across annotation paradigms. The zero-shot LLMs (baseline GPT-4o-mini and Llama-3.3-70B) inflate effect sizes roughly two- to threefold relative to human annotations, while fine-tuning attenuates them back toward the human scale. This magnitude profile is largely orthogonal to classification accuracy, so it is invisible to output-level comparison.

Our contributions are as follows:
\begin{enumerate}
    \item We propose a causal framework for testing annotation fidelity (comparing human and LLM annotators by holding sentiment constant and varying only ideology labels) that directly probes whether LLMs replicate the \textit{causal structure} of human judgments rather than merely their outputs.
    \item We demonstrate that the causal effect of topic sentiment on ideology replicates across human and LLM annotators for certain partisan topics, and that annotation paradigms diverge systematically in effect magnitude (zero-shot LLMs inflate it, fine-tuning attenuates it toward the human scale) and for others diverge in direction as well. This has implications for the safe use of LLM-generated silver labels in downstream analyses.
\end{enumerate}

\section{Related Work}

\paragraph{Sentiment and Ideology in Media.}
The relationship between sentiment, topic coverage, and political ideology has been studied across multiple domains. \citet{smirnova2017ideology} demonstrate that sentiment patterns in news coverage of specific topics correlate with ideological positioning, while \citet{bhatia2018topic} show that topic-specific sentiment analysis can help predict political ideology. \citet{bestvater2023sentiment} draw a distinction between sentiment and stance, arguing that target-aware classification is necessary for political text analysis. Beyond news media, similar sentiment-ideology relationships have been observed in congressional speech \citep{gentzkowshapiro2019congressionalspeech} and legislator tweets \citep{spell-etal-2020-embedding}. Our work differs from these studies in that we move beyond correlational analysis to estimate causal effects of topic sentiment on ideology, and we compare these effects across human and LLM annotation sources.

\paragraph{Causal Inference in NLP.}
Causal inference methods have been applied to text analysis across a range of settings. \citet{feder2022causalinferenceinnlp} provide a comprehensive survey of causal inference in NLP, covering settings where text serves as treatment, outcome, or confounder. \citet{veitch2019textembeddingscausalinference} propose using text embeddings to adjust for confounding in causal inference, while \citet{keith-etal-2021-text} develop a framework for text as causal mediator, estimating direct and indirect effects of social group signals through language. \citet{tierney-volfovsky-2021-sensitivity} apply causal mediation analysis to political polarization through text. In parallel, causal methods have been applied to understand LLM behavior, including gender bias detection via causal mediation analysis \citep{vig2020genderbiasLLMs}, mechanistic interpretation of arithmetic reasoning \citep{stolfo2023mechanisticinterpretationarithmeticreasoning}, and assessment of LLM comprehension \citep{han2024surfacestructurecausalassessment}. Our work contributes to this literature by applying causal inference not to explain model internals, but to diagnose systematic differences between human and LLM annotation behavior.

\paragraph{LLMs as Human Surrogates.}
A growing body of research explores whether LLMs can substitute for human participants in social science tasks \citep{horton2023large, li2024frontiers, ma-etal-2025-algorithmic, wu2023large, strachan2024testing}. Despite this promise, \citet{gao2025LLMscaution} provide a cautionary counterpoint: in an economic game requiring strategic reasoning, LLMs consistently fail to replicate the human behavior distribution, with failure modes that are inconsistent across models and input variations. These studies evaluate output-level agreement between LLMs and humans; our work complements them by asking whether the causal structure of LLM decisions, specifically which features drive predictions, mirrors that of human annotators.

\paragraph{Double Machine Learning.}
Our causal estimation relies on Double Machine Learning (DML), introduced by \citet{chernozhukov2018}, which provides root-N consistent estimates of treatment effects in the presence of high-dimensional confounders by using cross-fitting and Neyman-orthogonal score functions. DML is particularly suited to our setting because it accommodates the high-dimensional topic-presence confounders while allowing flexible first-stage models for the treatment and outcome nuisance functions.

\section{Experimental Setup}
\label{sec:methodology}

\subsection{Dataset and Expert Human Baseline}
AllSides tags both news sources and individual articles with categorical ideology ratings (Left, Center, Right); both kinds of labels are assigned through multiple rounds of review and consensus-building among a team of in-house experts\footnote{Details on the AllSides news curation policy can be found at \href{https://www.allsides.com/about/news-curation-principles}{https://www.allsides.com/about/news-curation-principles}.}. These AllSides article-level expert labels serve as our expert human baseline throughout the paper and the same labels are compared against each LLM paradigm under an identical experimental pipeline. Our article corpus builds on the AllSides corpus released by \citet{baly-etal-2020-detect} and is collected under the same news-curation protocol. To support reproducibility we release article-level metadata (source URL, outlet, ideology label, topic tags) together with the topic-sentiment annotations used in this paper, covering both the experimental subset and a broader collection of additional articles.\footnote{Available at \href{https://huggingface.co/datasets/upasanachatterjee/AllSides-sentiments}{https://huggingface.co/datasets/\\upasanachatterjee/AllSides-sentiments}} Our analyses operate on an $N=9{,}830$ set of expert-annotated articles.

\subsection{Causal Framework}
\begin{figure}[h]
    \centering
    \begin{tikzpicture}[->,shorten >=1pt,auto,node distance=3cm]
      \tikzstyle{every state}=[]

      \node(text) {$\text{Text } (X)$};
      \node(sentiment) [above of=text] {$\text{Sentiment } (F)$} edge [->] (text);
      \node(ideology) [below right of=sentiment] {$\text{Pred. Ideology } (Y)$} edge [<-] (sentiment) edge [<-] (text);
      \node(topic) [below left of=sentiment] {$\text{Topic } (T)$} edge [->] (text) edge [->, bend left] (ideology) edge [<->, dashed] (sentiment);
    \end{tikzpicture}
    \caption{Complete causal diagram for article ideology classification. Note, we do not include article text in our investigations.}
    \label{fig:causal_diagram}
\end{figure}
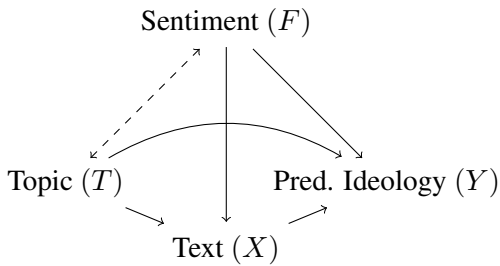

We model the relationships between article content, sentiment, and ideology using the causal diagram in Figure~\ref{fig:causal_diagram}, which formalizes our assumptions about the data-generating process and determines which variables must be controlled for to obtain unbiased causal estimates.

We define the following variables:
\begin{itemize}
    \item $X$: (text) the full article text.
    \item $T$: (topic tags) AllSides topic tags (e.g.\ Politics, Government Efficiency, Foreign Affairs).
    \item $F$: (sentiment) sentiment toward each topic, inferred from article text using Llama. Details are provided in Section~\ref{subsec:topic_sentiment_extraction}.
    \item $Y$: (perceived ideology) the ideology label assigned by the annotator (human and LLM).
\end{itemize}

Each arrow represents a hypothesized causal direction. The arrow from $X$ to $Y$ reflects that article content affects perceived ideology. The arrow from $F$ to $X$ reflects that sentiment toward a topic influences the written text. The bidirectional dashed arrow between $T$ and $F$ captures the mutual relationship between topics and sentiment: certain topics tend to evoke particular sentiments, while prevailing sentiment patterns influence which topics receive coverage.

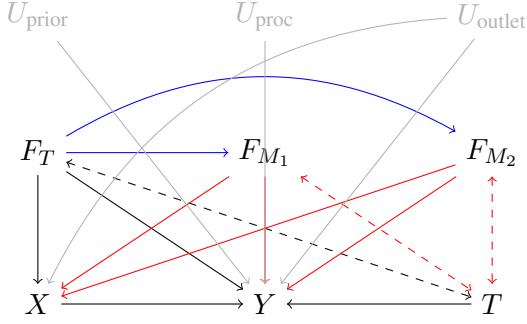
\begin{figure}[h!]
  \centering
  \begin{tikzpicture}[->,auto]
    \node[text=gray!60] (Uprior)  at (0,1.8)  {$U_{\text{prior}}$};
    \node[text=gray!60] (Uproc)   at (3,1.8)  {$U_{\text{proc}}$};
    \node[text=gray!60] (Uoutlet) at (6,1.8)  {$U_{\text{outlet}}$};

    \node (FT)   at (0,0)   {$F_{T}$};
    \node (FM1)  at (3,0)   {$F_{M_1}$};
    \node (FM2)  at (6,0)   {$F_{M_2}$};

    \node (Xabs) at (0,-2)  {$X$};
    \node (Y)    at (3,-2)  {$Y$};
    \node (T)    at (6,-2)  {$T$};

    \draw[->]           (FT) -- (Y);
    \draw[->]  (FT) -- (Xabs);
    \draw[<->, dashed]  (FT) -- (T);

    \draw[->]           (Xabs) -- (Y);
    \draw[->]  (T) to (Y);

    \draw[->, red]             (FM1) -- (Y);
    \draw[->, red]    (FM1) -- (Xabs);
    \draw[<->, dashed, red]    (FM1) -- (T);
    \draw[<-, blue]            (FM1) -- (FT);

    \draw[->, red]             (FM2) -- (Y);
    \draw[->, red]    (FM2) -- (Xabs);
    \draw[<->, dashed, red]    (FM2) -- (T);
    \draw[<-, blue, bend right] (FM2) to (FT);

    \draw[->, draw=gray!60]            (Uprior)  to (Y);
    \draw[->, draw=gray!60]            (Uproc)   -- (Y);
    \draw[->, draw=gray!60]            (Uoutlet) to (Y);
    \draw[->, draw=gray!60, bend right] (Uoutlet) to (Xabs);
  \end{tikzpicture}
  \caption{Expanded setup: sentiment variables as mediators. \textcolor{blue}{Blue} arrows show treatment-to-mediator paths; \textcolor{red}{red} shows mediator-to-outcome paths. \textcolor{gray}{Grey} nodes denote labeller-side factors absorbed into the per-annotator estimator and not modelled in this experiment.}
  \label{fig:causal3}
\end{figure}

In practice, each node expands into many variables. Figure~\ref{fig:causal3} illustrates this: each sentiment node $F$ corresponds to a distinct topic, and sentiment toward one topic ($F_T$, the treatment) may influence sentiment toward other topics ($F_{M_1}, F_{M_2}, \ldots$, the mediators), which in turn affect ideology classification $Y$. The structural relationships \textit{between} variable types remain as in Figure~\ref{fig:causal_diagram}; what changes is the dimensionality within each node.

The diagram represents a partial model of the labeling process. For human annotators, an abstraction of one cognitive channel (sentiment toward salient topics) used in arriving at an ideology label; for LLM annotators, it is an abstraction of one channel of the labeling pipeline. We apply the same SCM to both paradigms and compare them through this shared structural lens. That a single structure fits both is itself an assumption, and one of the things this paper effectively probes. Many additional features plausibly enter both judgments (illustrated in Figure~\ref{fig:causal3}); we deliberately scope to topic-level sentiment to keep the experimental design tractable and the estimands interpretable.

Applying a single SCM to both paradigms means we are comparing per-annotator conditional distributions $p(Y \mid X, \text{annotator})$ under a shared structural assumption rather than claiming the underlying processes are the same. Human annotators bring deliberation, prior beliefs, source/outlet priors, and between-annotator consensus dynamics that the diagram does not encode; LLM annotators carry training-data composition, instruction tuning, architecture, and hyperparameter choices that are likewise absent. Both sets of factors are absorbed into the per-annotator estimator for simplicity, at the cost of losing some nuance. As a result, divergences in estimated effects across annotators reflect the combined contribution of these unmodeled factors operating through the sentiment channel we do model. A more complete account would require separate causal probing of each annotator's data-generating process, which we treat as future work.

The roles of each variable in the estimation framework are:

\paragraph{Treatment Variable ($F_T$):} Sentiment toward a specific topic of interest (e.g., sentiment toward ``Immigration'' or ``Healthcare''). This is the variable whose causal effect on ideology we seek to estimate.

\paragraph{Outcome Variable ($Y$):} The ideology label (Left/Center/Right) assigned to the article.

\paragraph{Mediator Variables ($F_{M_1}, F_{M_2}, \ldots$):} Sentiment toward all other topics present in the article. These lie on the causal pathway between treatment sentiment and ideology: sentiment toward one topic may influence sentiment toward related topics, which in turn affects ideology classification. Blue arrows in Figure~\ref{fig:causal3} show these paths.

\paragraph{Confounder Variables:} $T$ (topic presence) is the primary confounder, since the presence of a topic influences both sentiment toward it and the overall ideological character of the article.

We additionally restrict the analysis to articles containing the treatment topic, preventing invalid comparisons between articles that discuss a topic and those that do not. This design supports two complementary families of causal estimands. The Average Treatment Effect (ATE) captures the overall causal impact of topic-community sentiment on ideology prediction, averaged across all articles in a community. Mediation analysis \citep{Pearl2001DirectAI} decomposes this total influence into a Natural Direct Effect (NDE), the portion of sentiment's impact that operates independently of other topic sentiments, and a Natural Indirect Effect (NIE), the portion that is mediated through co-occurring topic sentiments; the two sum to the Total Effect (TE). Together, these estimands allow us to ask not only whether sentiment toward a topic is causally associated with ideology prediction, but also through which pathways that association operates. Comparing human and LLM-derived estimates across both ATE and mediation quantities can reveal differences in how annotation sources represent these causal structures.

Two experimental notes that we want to make explicit:
\begin{itemize}
  \item $F$ is observational. The treatment is a sentiment summary statistic extracted from $X$, not a manipulation of the article itself. Identification leans on conditional ignorability given $T$ together with DML's nuisance modelling, not a natural-experiment design.
  \item Under greedy decoding, $Y$ is a deterministic function of $X$ for a fixed annotator, so reported variability is across-article rather than within-article.
\end{itemize}

\subsection{LLM Selection and Performance}
\begin{table}[h]
\centering
\begin{tabular}{@{}lc}
\toprule
 \textbf{LLM}& \textbf{F1-Macro} \\
 \midrule
    Finetuned GPT-4o-mini & 72.48 \\
    GPT-4o-mini & 50.07 \\
    Llama-3.3-70B-versatile & 54.61  \\
\bottomrule
\end{tabular}
\caption{Baseline LLM performance on AllSides political ideology classification.}
\label{tab:llm_baseline}
\end{table}

Table~\ref{tab:llm_baseline} shows the best model performance, evaluated against the expert human baseline as the ground truth. LLM evaluation setup details are provided in Appendix~\ref{sec:llm_evaluation}.

We tested a few different large language models from the GPT and Llama families on the ideology prediction task and selected the \texttt{gpt-4o-mini-2024-07-18} model snapshot for additional finetuning. This model was selected as it offered the most favorable balance between classification accuracy (as demonstrated in our baseline experiments) and computational cost among available OpenAI models. Details about the finetuning process and hyperparameters are provided in Section~\ref{sec:gpt_finetuning}.

\subsection{Topic Sentiment Extraction}
\label{subsec:topic_sentiment_extraction}
To extract sentiment variables ($F$), we apply a two-step process to the full article text ($X$): first, named entities and key concepts are extracted and assigned sentiment polarity on a continuous scale from $-1.0$ (strongly negative) to $+1.0$ (strongly positive); second, each entity is associated with predefined topic tags assigned by the AllSides team, and entity sentiment scores are aggregated by topic to produce a per-topic sentiment profile for each article. Full model and implementation details are provided in Appendix~\ref{sec:topic_sentiment_extraction}.

\subsection{Community Detection}
To identify the topical structure of the article corpus, we apply the Louvain community detection algorithm to a co-occurrence graph of topic tags, where nodes are tags and edge weights reflect how often two tags appear together in the same article. Louvain is well-suited to this setting because it maximises modularity directly on the co-occurrence graph, grouping topics that frequently co-appear into the same community. Since co-occurrence is a direct measure of how strongly two topic treatment columns move together across articles, the resulting communities correspond to natural clusters of correlated treatments. Graph construction details are provided in Appendix~\ref{sec:community_detection}.

For our community-level multi-treatment DML experiments, we limited our analysis to communities where each topic has at least 5 articles, leaving nine communities.

\subsection{Experimental Settings}
We use \texttt{LinearDML} \citep{chernozhukov2018} with PCA-reduced confounders and continuous sentiment scores, with bootstrap confidence intervals ($B=2000$). Full experimental settings are provided in Appendix~\ref{sec:community_detection}.

\section{Experiments and Results}

All experiments are conducted across four annotation paradigms: the LLMs introduced in Table~\ref{tab:llm_baseline} and expert human annotators. This multi-way comparison lets us examine whether causal relationships are stable across annotation approaches or vary systematically between human judgment and large language models\footnote{Code available at \href{https://github.com/upasanachatterjee/causal-inference-on-text}{https://github.com/upasanachatterjee/ causal-inference-on-text}}.

\paragraph{Estimation Details.} The outcome variable $Y$ is encoded as an ordinal (Left${}=0$, Center${}=1$, Right${}=2$) and treated as continuous, with \texttt{RandomForestRegressor} as the nuisance model for both outcome and treatment. This encoding assumes equal spacing between adjacent ideology categories: a unit ATE of $+1.0$ corresponds to a one-category rightward shift in expected ideology (e.g., ATE $= -0.036$ represents a 3.6\% shift toward Left). EconML's \texttt{LinearDML} handles cross-fitting of nuisance models internally. Significance is determined by whether the 95\% confidence interval (CI) excludes zero, equivalent to a two-sided test at $\alpha = 0.05$; intervals are non-parametric bootstrap percentile CIs ($B = 2{,}000$).

We present results at two levels of granularity. We lead with the \emph{topic level}, where each treatment is sentiment toward a single AllSides topic tag, because this is where the causal signal is strongest and most interpretable. We then aggregate topics into detected communities; those coarser \emph{community-level} results are consistent with the topic level and are deferred to in Appendix~\ref{sec:community_ate}.

\paragraph{Multi-treatment ATE.} We fit a single \texttt{LinearDML} model per paradigm with the top 20 topics (by article count) as simultaneous treatments, controlling for topic-presence confounders via PCA-compressed $W$. Each coefficient is the ATE of a one-standard-deviation increase in that topic's sentiment, holding all other topic sentiments fixed. Of the 20 topics, nine produce at least one significant effect across the four paradigms; Figure~\ref{fig:multi-continuous} shows these nine. The full 20-topic table with 95\% CIs is in Appendix~\ref{sec:topic_level_results}.

\begin{figure}[t]
    \centering
    \includegraphics[width=\linewidth]{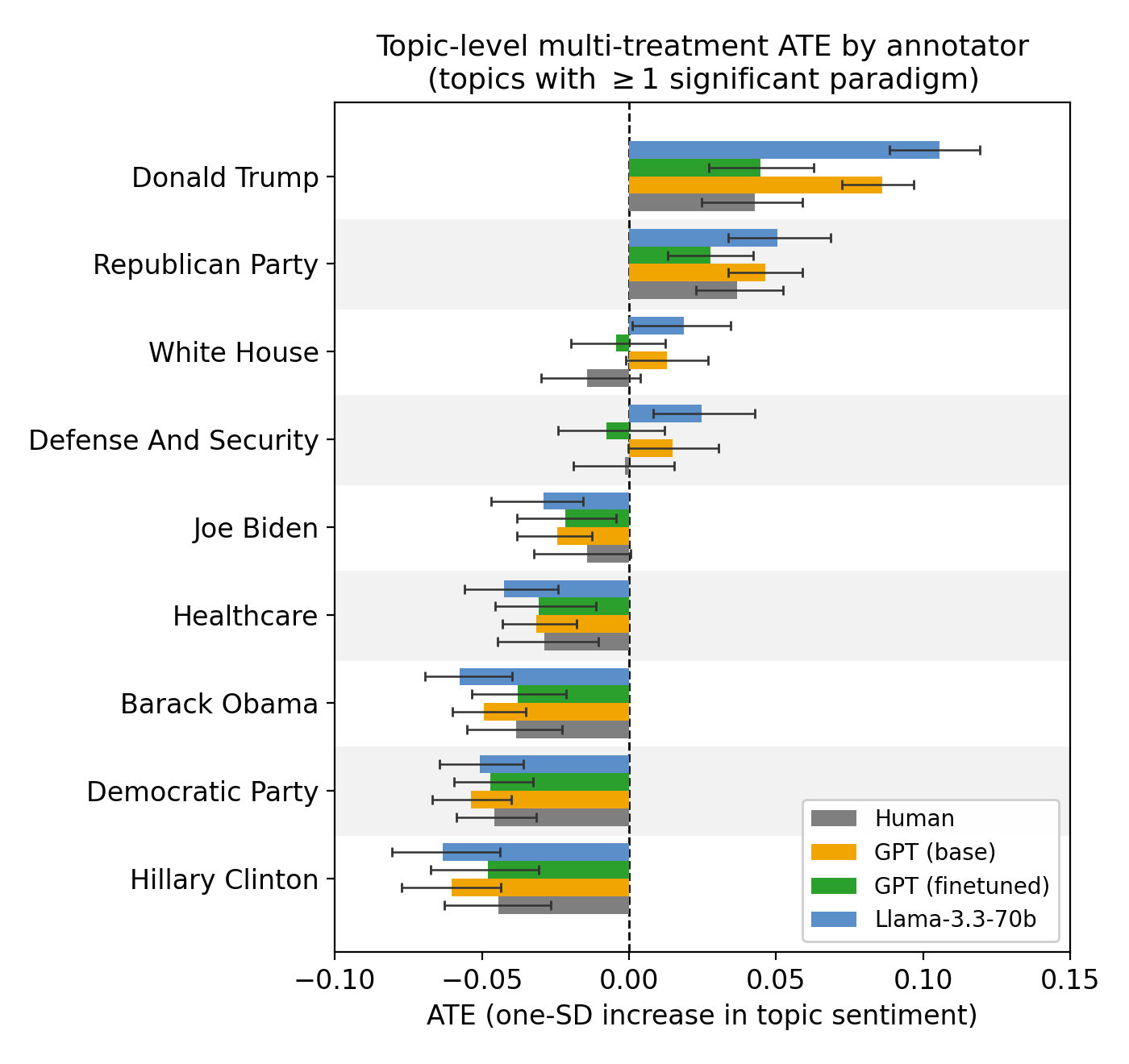}
    \caption{Multi-treatment DML: ATE per topic (one-SD increase in sentiment) for the nine of 20 qualifying topics with at least one significant paradigm. Error bars are 95\% bootstrap CIs; an effect is significant when its interval excludes zero. Full table with all 20 topics and CIs in Appendix~\ref{sec:topic_level_results}, Table~\ref{tab:multi-continuous-full}.}
    \label{fig:multi-continuous}
\end{figure}

The directional structure is interpretable and largely partisan: topics associated with Republican figures (Donald Trump, Republican Party) carry positive ATEs (more positive sentiment $\to$ rightward shift), while topics associated with Democratic figures and left-coded policy (Barack Obama, Joe Biden, Hillary Clinton, Democratic Party, Healthcare) carry negative ATEs. In particular, six topics---Donald Trump, Republican Party, Healthcare, Barack Obama, Hillary Clinton, and Democratic Party---are significant under all four paradigms including expert humans. Joe Biden is significant under three (all but Human). The replication of this relationship across multiple annotation paradigms suggests that the sentiment–ideology association reflects systematic partisan structure in annotator priors rather than an artifact of any single annotation paradigm.

Two further patterns stand out. First, significance tracks how polarising a topic is, not its article count: the three largest topics (Elections $N{=}1709$, Politics $N{=}1520$, Presidential Elections $N{=}1331$; Appendix~\ref{sec:topic_level_results}) produce no significant effects under any paradigm, while smaller partisan topics such as Democratic Party ($N{=}261$) are significant everywhere. Second, directional agreement is not a general guarantee. The topics White House and Defense and Security reach significance under only Llama, with a divergent sign from other paradigms (Llama predicts left while the others predict right). On topics where no annotator reaches significance, such as Politics and Russia, the disagreement is broader: even the baseline and fine-tuned GPT variants, which share an architecture and differ only in fine-tuning, point in opposite directions. This could be due to partisan shift in the corpus over time, and underscores the importance of future work (see Section~\ref{sec:future_work}) using a more comprehensive SCM with richer features.

\begin{figure}[t]
    \centering
    \includegraphics[width=\linewidth]{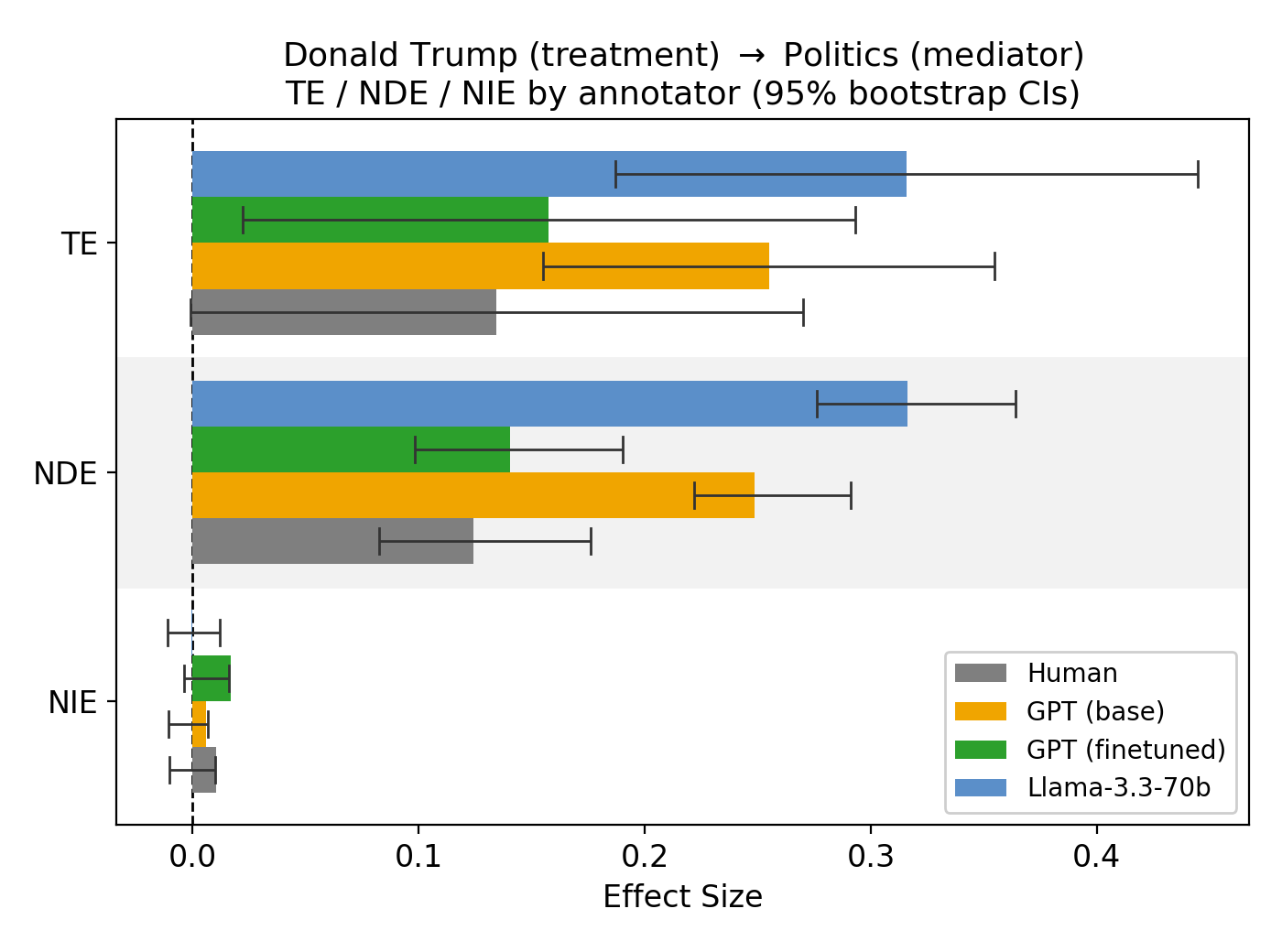}
    \caption{Mediation decomposition for Donald Trump (treatment) $\rightarrow$ Politics (mediator), $Q_{25}\to Q_{75}$ contrast, $N=1{,}424$. Error bars are 95\% bootstrap CIs for TE, NDE, and NIE.}
    \label{fig:trump_mediation}
\end{figure}

\paragraph{Donald Trump $\rightarrow$ Politics mediation.} We decompose the strongest topic effect into direct and indirect channels, taking Donald Trump sentiment as treatment and Politics sentiment as mediator ($N=1{,}424$ articles). Figure~\ref{fig:trump_mediation} reports the Total Effect (TE), Natural Direct Effect (NDE), and Natural Indirect Effect (NIE) for the $Q_{25}\to Q_{75}$ contrast; full numbers, are in Appendix~\ref{sec:community_mediation_results}, Table~\ref{tab:te-trump}.

All four paradigms agree that more positive Trump sentiment shifts ideology rightward, and three of four reach significance; only Human narrowly misses ($\widehat{\text{TE}} = +0.135$, CI $[-0.001, +0.270]$). The direct channel dominates in every paradigm ($|\text{NDE}| \gg |\text{NIE}|$), so the Trump-sentiment effect operates almost entirely independently of Politics sentiment. The magnitude ordering is telling: the zero-shot LLMs produce the largest effects (Llama $+0.316$, GPT base $+0.255$), while fine-tuned GPT ($+0.158$) sits closest to the human estimate. Fine-tuning attenuates the signal relative to the zero-shot GPT baseline rather than amplifying it.

\paragraph{Community-level results.} Aggregating topics into the nine Louvain communities that pass the minimum-presence threshold (Section~\ref{sec:methodology}) yields effects that are consistent with the topic level but weaker, since pooling dilutes any single partisan signal. The headline is that Community~2 (Public Health) replicates across all four paradigms (significant negative ATEs), the community-level analogue of the fully replicated partisan topics; community-level mediation is again concentrated in the direct channel (NDE) rather than the indirect one. Full community-level ATE and mediation results are reported in Appendices~\ref{sec:community_ate} and~\ref{sec:community_mediation_results}.

\section{Discussion}

Our results show that topic sentiment carries a genuine causal relationship with ideology classification, and that this relationship is largely shared across annotation paradigms. As only the ideology labels vary across the four datasets, the remaining differences across paradigms reflect systematic differences in how each labelling source assigns ideological stance to article text.

\subsection{Polarisation Drives the Causal Signal}

Effect significance tracks how polarising a topic is, not how frequently it appears---but ``polarising'' is a property of the \emph{annotator}: our estimates recover how polarising each source (Human, GPT, Llama) \emph{perceives} a topic to be, a perception that encodes priors---belief-based for the human experts, training-based for the models. The three largest topics by article count (Elections, Politics, Presidential Elections) produce no significant effects under any paradigm, while smaller but explicitly partisan topics (Democratic Party, Hillary Clinton, Republican Party) are significant across paradigms. 

The directional structure of the significant effects is consistent with partisan expectations: sentiment toward Democratic figures and left-coded policy (Barack Obama, Joe Biden, Hillary Clinton, Democratic Party, Healthcare) shifts predictions leftward, while sentiment toward Republican figures (Donald Trump, Republican Party) shifts them rightward. This alignment holds across all four paradigms and across both analysis levels, with Community~2 (Public Health) reproducing full cross-paradigm replication at the community level. That all four annotators exhibit the same partisan structure indicates a shared prior about which topics are polarising and in which direction, rather than a coupling specific to any single source---consistent with work showing that human annotators draw on metalinguistic framing cues, including sentiment toward salient entities, when assigning ideological stance \citep{nagy2007metalinguistic, entman2010mediaframingbiases, smirnova2017ideology}.

\subsection{Classification Accuracy Does Not Predict Causal Fidelity}

The magnitude comparison is largely orthogonal to classification accuracy. Fine-tuned GPT-4o-mini achieves the highest F1 among the machine annotators (72.48, Table~\ref{tab:llm_baseline}), far above Llama-3.3-70B (54.61), yet Llama's causal profile is the most inflated and fine-tuned GPT's is the closest to human. F1 measures whether a model assigns the correct label; the causal analysis measures which textual features drive that assignment. High performance on one axis does not guarantee alignment on the other, echoing arguments that causal interpretability is necessary precisely because correlation-based evaluation leaves spurious feature reliance invisible \citep{moraffah2020causal}.

\subsection{Paradigms Largely Diverge in Magnitude}

Where paradigms largely differ is in the magnitude of the estimated effects. Across the topics with cross-paradigm significance, the zero-shot LLMs (Llama-3.3-70B and GPT-4o-mini baseline) consistently produce the largest ATEs, while human annotations produce the smallest; the fine-tuned models fall in between and closest to the human scale. The Donald Trump pathway makes this concrete: using Llama labels a researcher would estimate an ATE of $+0.106$, and using human labels $+0.043$---roughly $2.5\times$ smaller---while fine-tuned GPT ($+0.045$) falls within $5\%$ of the human estimate. Both are significant and directionally identical, but the magnitude gap could support different conclusions about the effect's practical importance. 

In the mediation decomposition the same ordering holds: fine-tuned GPT's Total Effect ($+0.158$) brackets the human estimate ($+0.135$), while the zero-shot LLMs inflate it two- to three-fold. Fine-tuning therefore acts as a magnitude attenuator rather than a directional corrector: it does not change which topics are significant or which direction effects point, but it compresses inflated zero-shot effects toward the human range. These magnitude gaps are a symptom of divergent priors: each scheme has internalised a different sense of how much sentiment toward a topic should move an ideology judgment, with the zero-shot models holding a systematically stronger such prior than humans. Fixed by training rather than by the corpus, this is the component of annotator behaviour most likely to misfire under distribution shift, which the downstream risk we turn to below.

We can also see the effect of divergent priors in in the case of the Politics and Russia topics. Both are non-significant under all paradigms, but have differently signed estimates under the zero-shot GPT and fine-tuned GPT paradigms. In fact, the human annotators agree with fine-tuned GPT on the sign of the Politics effect, but with zero-shot GPT on the sign of the Russia effect (see Table~\ref{tab:multi-continuous-full}). So the choice of annotator could determine the sign of a reported (if non-significant) relationship. This directional instability reinforces the need for multi-paradigm validation when LLM annotations feed causal pipelines.

\subsection{A Localised Shortcut and Its Limits}

One place a machine paradigm detects an aggregate effect that human labels do not is Community~1 (Elections and Campaigns): the two zero-shot paradigms (GPT baseline $+0.034$, Llama $+0.056$) reach significance while Human ($+0.016$) and fine-tuned GPT ($+0.009$) do not. This is consistent with a localised shortcut, in which zero-shot models respond to surface-level cues associated with electoral content that spuriously correlate with ideology, and fine-tuning on human labels compresses that reliance back toward the human null. This aligns with work showing that shortcut reliance persists in zero-shot and few-shot LLM settings and that supervised adaptation helps mitigate it \citep{yuan2024llmsovercomeshortcutlearning, zhou2024explorespuriouscorrelationsconcept}. The effect is community-specific rather than pervasive, however: on the strongly partisan topics every paradigm agrees, so the shortcut manifests only where the underlying partisan signal is weak.

\subsection{Implications for LLM Annotation}

A growing line of work explores LLMs as replacements for human annotators and study participants \citep{horton2023large, ma-etal-2025-algorithmic, gao2025LLMscaution}. Our findings give a nuanced picture. On the positive side, for the partisan topics where all four annotators reach significance, they agree on direction with no reversals; for detecting the presence and direction of a sentiment--ideology relationship on such topics, LLM annotations are reliable proxies for human labels. But this agreement is scoped to those fully-significant topics rather than being a general property. The magnitude of estimated effects varies substantially even where the sign agrees: zero-shot labels systematically overstate effect sizes, and if used as silver-standard data or as inputs to causal analyses this inflation propagates. Fine-tuned models mitigate but do not eliminate the gap.

The practical stakes are clearest out of domain. The sentiment effects are small, but the consistent inflation by the zero-shot models is not benign: it encodes a prior---that sentiment toward a topic strongly signals ideology---which held as those topics were polarised during training but need not persist. A model that over-learns this coupling will misclassify precisely where it weakens: on a topic that de-polarises over time, or on a complex, balanced article whose sentiment cues cut against its stance. Fine-tuning on human labels tempers the prior, which is why the fine-tuned estimates track the human scale, whereas the un-finetuned models risk systematic error on inputs far from their training distribution. Because such divergent priors are invisible to in-distribution accuracy, annotators that agree on today's benchmark can fail in opposite directions once the distribution shifts---a catastrophic downstream failure that only a causal, prior-aware audit surfaces.

\subsection{Mediation Analysis as a Diagnostic Framework}

Output-level metrics do not expose why a model produces a given label, and LLMs cannot be relied upon to self-report the causal basis of their annotations: they perform near-randomly on pure causal inference from correlational evidence \citep{jin2024largelanguagemodelsinfer} and are prone to post hoc and ordering fallacies when reasoning about causation \citep{joshi-etal-2024-llms}. External causal analysis is therefore required. Our mediation framework provides one such approach: by holding the input text and causal structure fixed and varying only the annotation paradigm, we isolate differences attributable to the annotator rather than the data, consistent with calls for structured causal evaluation pipelines beyond accuracy benchmarks \citep{cheng2022evaluation}. Applied to annotation auditing, it surfaces which textual features drive predictions through direct versus mediated pathways, offering a basis for trustworthiness assessment that F1 and agreement metrics cannot provide.

\section{Future Work}
\label{sec:future_work}

Our framework treats each topic sentiment as a separate treatment or mediator. Modeling multiple topic sentiments as co-treatments would capture cross-topic interactions, and incorporating the article-level features discussed in our limitations (outlet, author, date, length, lexical complexity, framing) as additional treatments, mediators, or confounders would yield a richer multi-feature SCM that better isolates which textual dimensions drive the human-LLM divergences we observe.

A complementary direction is to probe each annotator's data-generating process separately rather than under a shared SCM: a human-specific diagram would represent deliberation, prior beliefs, and consensus dynamics, while an LLM-specific diagram would represent training-data and fine-tuning effects. Comparing the two paradigm-specific causal profiles would more directly characterize where and why human and LLM annotations diverge.

Our mediation-as-diagnostic framework could be applied beyond political ideology to other annotation tasks where LLM labels are used as silver-standard training data, such as sentiment analysis, stance detection, or toxicity classification. Comparing mediation pathways across human and LLM annotators in these domains would help determine whether the sentiment-sensitivity patterns we find are specific to ideology or reflect a more general tendency of LLMs.

Our current study is limited to two LLM families (GPT-4o-mini and Llama-3.3-70B). Expanding the investigation to a broader set of models, particularly those with reported ideological tendencies such as Grok \citep{xai2024grok}, would allow a more systematic characterisation of how baseline ideological priors interact with fine-tuning to shape causal pathways. A model with a reported rightward lean, for instance, may exhibit sign reversals in different community configurations than a centrist baseline model, or may differ in its susceptibility to acquiring spurious sentiment-ideology couplings during fine-tuning. Comparing causal profiles across a wider model zoo would help disentangle the contributions of pre-training ideology, instruction tuning, and task-specific fine-tuning to the divergences we observe.

\section{Limitations}

\paragraph{Sentiment is operationally an LLM construct.}
Throughout the paper, what we call ``sentiment'' ($F$) is operationally Llama-3.3-70B's sentiment estimate, not a human-validated construct: we did not human-annotate any subset of the extractions. The treatment variable should therefore be read as ``Llama-extracted sentiment'' rather than ``sentiment, full stop,'' and the entire analysis is conditional on sentiment that is legible to Llama-3.3-70B. Cases where a human reader would perceive sentiment but the extractor returns nothing (or misses the entity-topic association) are systematically absent from $F$, biasing estimates toward LLM-legible features and likely understating the role of more implicit or contextual cues that human annotators could still use. 

Beyond presence/absence, both the direction and magnitude of an extracted sentiment score are annotator-dependent artifacts. Direction (positive vs.\ negative) is plausibly more stable across annotators, but magnitude can vary substantially between annotators (human or model) even when they agree on sign. Because our DML and mediation estimates are driven by variation in the magnitude of $F$, annotator-specific scale and calibration differences propagate directly into the estimated effects.

\paragraph{Unmodeled article-level features.}
Our causal framework controls for topic presence but does not encode outlet identity, author, publication date, article length, or lexical complexity. Outlets in particular may carry systematic sentiment patterns that confound the sentiment-ideology relationship; controlling for outlet identity would address this but would also substantially reduce the variation available for estimation given the correlation between outlet and ideology.

\paragraph{Statistical power.}
While the multi-treatment ATEs are estimated on the full $N=9{,}830$ corpus, the mediation analyses restrict to the subpopulation where the treatment (and mediator) are present, so some pathways operate on considerably smaller effective samples (e.g.\ $N=1{,}116$ for the Community~$2\to1$ pathway). This limits power for the mediation analyses in particular and produces the wide Total-Effect confidence intervals reported in Appendix~\ref{sec:community_mediation_results}. Non-significant estimates (such as the Human Total Effect on the Donald Trump pathway, which narrowly misses) should accordingly be interpreted with caution, as they may reflect insufficient power rather than a true absence of effects.

\paragraph{Bounding the shared-SCM claim.}
Section~\ref{sec:methodology} frames the same SCM as a deliberately simplified, sentiment-only model of the labeling process for both human and LLM annotators, with per-annotator factors absorbed into the per-annotator estimator. Reported effects should accordingly be read as effects propagating through the topic-level sentiment channel, conditional on the topics actually present in each article, rather than as claims about the underlying generative processes of human or LLM judgment.


\bibliography{custom}

@inproceedings{bhatia2018topic,
  title={Topic-specific sentiment analysis can help identify political ideology},
  author={Bhatia, Sumit and Deepak, P},
  booktitle={Proceedings of the 9th workshop on computational approaches to subjectivity, sentiment and social media analysis},
  pages={79--84},
  year={2018}
}

@article{bestvater2023sentiment,
  title={Sentiment is not stance: Target-aware opinion classification for political text analysis},
  author={Bestvater, Samuel E and Monroe, Burt L},
  journal={Political Analysis},
  volume={31},
  number={2},
  pages={235--256},
  year={2023},
  publisher={Cambridge University Press}
}

@article{smirnova2017ideology,
  title={Ideology through sentiment analysis: A changing perspective on Russia and Islam in NYT},
  author={Smirnova, Anastasia and Laranetto, Helena and Kolenda, Nicholas},
  journal={Discourse \& Communication},
  volume={11},
  number={3},
  pages={296--313},
  year={2017},
  publisher={SAGE Publications Sage UK: London, England}
}

@article{chernozhukov2018,
  author  = {Chernozhukov, Victor and Chetverikov, Denis and Demirer, Mert and
             Duflo, Esther and Hansen, Christian and Newey, Whitney and Robins, James},
  title   = {Double/debiased machine learning for treatment and structural parameters},
  journal = {The Econometrics Journal},
  year    = {2018},
  volume  = {21},
  number  = {1},
  pages   = {C1--C68},
}

@inproceedings{tierney-volfovsky-2021-sensitivity,
    title = "Sensitivity Analysis for Causal Mediation through Text: an Application to Political Polarization",
    author = "Tierney, Graham  and
      Volfovsky, Alexander",
    editor = "Feder, Amir  and
      Keith, Katherine  and
      Manzoor, Emaad  and
      Pryzant, Reid  and
      Sridhar, Dhanya  and
      Wood-Doughty, Zach  and
      Eisenstein, Jacob  and
      Grimmer, Justin  and
      Reichart, Roi  and
      Roberts, Molly  and
      Shalit, Uri  and
      Stewart, Brandon  and
      Veitch, Victor  and
      Yang, Diyi",
    booktitle = "Proceedings of the First Workshop on Causal Inference and NLP",
    month = nov,
    year = "2021",
    address = "Punta Cana, Dominican Republic",
    publisher = "Association for Computational Linguistics",
    url = "https://aclanthology.org/2021.cinlp-1.5/",
    doi = "10.18653/v1/2021.cinlp-1.5",
    pages = "61--73",
}

@article{gentzkowshapiro2019congressionalspeech,
author = {Gentzkow, Matthew and Shapiro, Jesse M. and Taddy, Matt},
title = {Measuring Group Differences in High-Dimensional Choices: Method and Application to Congressional Speech},
journal = {Econometrica},
volume = {87},
number = {4},
pages = {1307-1340},
doi = {https://doi.org/10.3982/ECTA16566},
url = {https://onlinelibrary.wiley.com/doi/abs/10.3982/ECTA16566},
eprint = {https://onlinelibrary.wiley.com/doi/pdf/10.3982/ECTA16566},
year = {2019}
}

@inproceedings{spell-etal-2020-embedding,
    title = "An {E}mbedding {M}odel for {E}stimating {L}egislative {P}references from the {F}requency and {S}entiment of {T}weets",
    author = "Spell, Gregory  and
      Guay, Brian  and
      Hillygus, Sunshine  and
      Carin, Lawrence",
    editor = "Webber, Bonnie  and
      Cohn, Trevor  and
      He, Yulan  and
      Liu, Yang",
    booktitle = "Proceedings of the 2020 Conference on Empirical Methods in Natural Language Processing (EMNLP)",
    month = nov,
    year = "2020",
    address = "Online",
    publisher = "Association for Computational Linguistics",
    url = "https://aclanthology.org/2020.emnlp-main.46/",
    doi = "10.18653/v1/2020.emnlp-main.46",
    pages = "627--641"
}

@inproceedings{keith-etal-2021-text,
    title = "Text as Causal Mediators: Research Design for Causal Estimates of Differential Treatment of Social Groups via Language Aspects",
    author = "Keith, Katherine  and
      Rice, Douglas  and
      O{'}Connor, Brendan",
    editor = "Feder, Amir  and
      Keith, Katherine  and
      Manzoor, Emaad  and
      Pryzant, Reid  and
      Sridhar, Dhanya  and
      Wood-Doughty, Zach  and
      Eisenstein, Jacob  and
      Grimmer, Justin  and
      Reichart, Roi  and
      Roberts, Molly  and
      Shalit, Uri  and
      Stewart, Brandon  and
      Veitch, Victor  and
      Yang, Diyi",
    booktitle = "Proceedings of the First Workshop on Causal Inference and NLP",
    month = nov,
    year = "2021",
    address = "Punta Cana, Dominican Republic",
    publisher = "Association for Computational Linguistics",
    url = "https://aclanthology.org/2021.cinlp-1.2/",
    doi = "10.18653/v1/2021.cinlp-1.2",
    pages = "21--32"
}

@inproceedings{veitch2019textembeddingscausalinference,
  title={Adapting text embeddings for causal inference},
  author={Veitch, Victor and Sridhar, Dhanya and Blei, David},
  booktitle={Conference on uncertainty in artificial intelligence},
  pages={919--928},
  year={2020},
  organization={PMLR}
}

@inproceedings{vig2020genderbiasLLMs,
 author = {Vig, Jesse and Gehrmann, Sebastian and Belinkov, Yonatan and Qian, Sharon and Nevo, Daniel and Singer, Yaron and Shieber, Stuart},
 booktitle = {Advances in Neural Information Processing Systems},
 editor = {H. Larochelle and M. Ranzato and R. Hadsell and M.F. Balcan and H. Lin},
 pages = {12388--12401},
 publisher = {Curran Associates, Inc.},
 title = {Investigating Gender Bias in Language Models Using Causal Mediation Analysis},
 url = {https://proceedings.neurips.cc/paper_files/paper/2020/file/92650b2e92217715fe312e6fa7b90d82-Paper.pdf},
 volume = {33},
 year = {2020}
}

@inproceedings{stolfo2023mechanisticinterpretationarithmeticreasoning,
    title = "A Mechanistic Interpretation of Arithmetic Reasoning in Language Models using Causal Mediation Analysis",
    author = "Stolfo, Alessandro  and
      Belinkov, Yonatan  and
      Sachan, Mrinmaya",
    editor = "Bouamor, Houda  and
      Pino, Juan  and
      Bali, Kalika",
    booktitle = "Proceedings of the 2023 Conference on Empirical Methods in Natural Language Processing",
    month = dec,
    year = "2023",
    address = "Singapore",
    publisher = "Association for Computational Linguistics",
    url = "https://aclanthology.org/2023.emnlp-main.435/",
    doi = "10.18653/v1/2023.emnlp-main.435",
    pages = "7035--7052"
}

@inproceedings{
han2024surfacestructurecausalassessment,
title={Beyond Surface Structure: A Causal Assessment of {LLM}s' Comprehension ability},
author={Yujin Han and Lei Xu and Sirui Chen and Difan Zou and Chaochao Lu},
booktitle={The Thirteenth International Conference on Learning Representations},
year={2025},
url={https://openreview.net/forum?id=gsShHPxkUW}
}

@article{feder2022causalinferenceinnlp,
  author = {Amir Feder and Katherine A. Keith and Emaad Manzoor and Reid Pryzant and Dhanya Sridhar and Zach Wood-Doughty and Jacob Eisenstein and Justin Grimmer and Roi Reichart and Margaret E. Roberts and Brandon M. Stewart and Victor Veitch and Diyi Yang},
  title = {Causal Inference in Natural Language Processing: Estimation, Prediction, Interpretation and Beyond},
  year = {2022},
  journal = {Transactions of the Association for Computational Linguistics},
  volume = {10},
  url = {https://transacl.org/index.php/tacl/article/view/4005},
  language = {eng},
}

@article{Pearl2001DirectAI,
  title={Direct and Indirect Effects},
  author={Judea Pearl},
  journal={Probabilistic and Causal Inference},
  year={2001},
  url={https://api.semanticscholar.org/CorpusID:5947965}
}

@inproceedings{baly-etal-2020-detect,
    title = "We Can Detect Your Bias: Predicting the Political Ideology of News Articles",
    author = "Baly, Ramy  and
      Da San Martino, Giovanni  and
      Glass, James  and
      Nakov, Preslav",
    editor = "Webber, Bonnie  and
      Cohn, Trevor  and
      He, Yulan  and
      Liu, Yang",
    booktitle = "Proceedings of the 2020 Conference on Empirical Methods in Natural Language Processing (EMNLP)",
    month = nov,
    year = "2020",
    address = "Online",
    publisher = "Association for Computational Linguistics",
    url = "https://aclanthology.org/2020.emnlp-main.404/",
    doi = "10.18653/v1/2020.emnlp-main.404",
    pages = "4982--4991"
}

@article{entman2010mediaframingbiases,
author = {Robert M. Entman},
title ={Media framing biases and political power: Explaining slant in news of Campaign 2008},
journal = {Journalism},
volume = {11},
number = {4},
pages = {389-408},
year = {2010},
doi = {10.1177/1464884910367587},
URL = {https://doi.org/10.1177/1464884910367587},
eprint = { https://doi.org/10.1177/1464884910367587
}}

@inproceedings{ma-etal-2025-algorithmic,
    title = "Algorithmic Fidelity of Large Language Models in Generating Synthetic {G}erman Public Opinions: A Case Study",
    author = "Ma, Bolei  and
      Yoztyurk, Berk  and
      Haensch, Anna-Carolina  and
      Wang, Xinpeng  and
      Herklotz, Markus  and
      Kreuter, Frauke  and
      Plank, Barbara  and
      A{\ss}enmacher, Matthias",
    editor = "Che, Wanxiang  and
      Nabende, Joyce  and
      Shutova, Ekaterina  and
      Pilehvar, Mohammad Taher",
    booktitle = "Proceedings of the 63rd Annual Meeting of the Association for Computational Linguistics (Volume 1: Long Papers)",
    month = jul,
    year = "2025",
    address = "Vienna, Austria",
    publisher = "Association for Computational Linguistics",
    url = "https://aclanthology.org/2025.acl-long.90/",
    doi = "10.18653/v1/2025.acl-long.90",
    pages = "1785--1809",
    ISBN = "979-8-89176-251-0"
}

@article{wu2023large,
  title={Large language models can be used to estimate the latent positions of politicians},
  author={Wu, Patrick Y and Nagler, Jonathan and Tucker, Joshua A and Messing, Solomon},
  url={https://api.semanticscholar.org/CorpusID:257636745},
  year={2023}
}

@article{strachan2024testing,
  title={Testing theory of mind in large language models and humans},
  author={Strachan, James WA and Albergo, Dalila and Borghini, Giulia and Pansardi, Oriana and Scaliti, Eugenio and Gupta, Saurabh and Saxena, Krati and Rufo, Alessandro and Panzeri, Stefano and Manzi, Guido and others},
  journal={Nature human behaviour},
  volume={8},
  number={7},
  pages={1285--1295},
  year={2024},
  publisher={Nature Publishing Group UK London}
}

@article{li2024frontiers,
  title={Frontiers: Determining the validity of large language models for automated perceptual analysis},
  author={Li, Peiyao and Castelo, Noah and Katona, Zsolt and Sarvary, Miklos},
  journal={Marketing Science},
  volume={43},
  number={2},
  pages={254--266},
  year={2024},
  publisher={INFORMS}
}

@techreport{horton2023large,
  title={Large language models as simulated economic agents: What can we learn from homo silicus?},
  author={Horton, John J},
  year={2023},
  institution={National Bureau of Economic Research}
}

@article{gao2025LLMscaution,
author = {Yuan Gao  and Dokyun Lee  and Gordon Burtch  and Sina Fazelpour },
title = {Take caution in using LLMs as human surrogates},
journal = {Proceedings of the National Academy of Sciences},
volume = {122},
number = {24},
pages = {e2501660122},
year = {2025},
doi = {10.1073/pnas.2501660122},
URL = {https://www.pnas.org/doi/abs/10.1073/pnas.2501660122},
eprint = {https://www.pnas.org/doi/pdf/10.1073/pnas.2501660122}}

@article{nagy2007metalinguistic,
  title={Metalinguistic awareness and the vocabulary-comprehension connection},
  author={Nagy, William},
  journal={Vocabulary acquisition: Implications for reading comprehension},
  pages={52--77},
  year={2007},
  publisher={Guildford}
}

@inproceedings{yuan2024llmsovercomeshortcutlearning,
  title={Do llms overcome shortcut learning? an evaluation of shortcut challenges in large language models},
  author={Yuan, Yu and Zhao, Lili and Zhang, Kai and Zheng, Guangting and Liu, Qi},
  booktitle={Proceedings of the 2024 Conference on Empirical Methods in Natural Language Processing},
  pages={12188--12200},
  year={2024}
}

@inproceedings{zhou2024explorespuriouscorrelationsconcept,
  title={Explore spurious correlations at the concept level in language models for text classification},
  author={Zhou, Yuhang and Xu, Paiheng and Liu, Xiaoyu and An, Bang and Ai, Wei and Huang, Furong},
  booktitle={Proceedings of the 62nd Annual Meeting of the Association for Computational Linguistics (Volume 1: Long Papers)},
  pages={478--492},
  year={2024}
}

@misc{xai2024grok,
  title={{Grok}: A Large Language Model by {xAI}},
  author={{xAI}},
  year={2025},
  url={https://data.x.ai/2025-08-20-grok-4-model-card.pdf},
}

@inproceedings{joshi-etal-2024-llms,
    title = "{LLM}s Are Prone to Fallacies in Causal Inference",
    author = "Joshi, Nitish  and
      Saparov, Abulhair  and
      Wang, Yixin  and
      He, He",
    editor = "Al-Onaizan, Yaser  and
      Bansal, Mohit  and
      Chen, Yun-Nung",
    booktitle = "Proceedings of the 2024 Conference on Empirical Methods in Natural Language Processing",
    month = nov,
    year = "2024",
    address = "Miami, Florida, USA",
    publisher = "Association for Computational Linguistics",
    url = "https://aclanthology.org/2024.emnlp-main.590/",
    doi = "10.18653/v1/2024.emnlp-main.590",
    pages = "10553--10569"
}

@inproceedings{jin2024largelanguagemodelsinfer,
  title     = {Can Large Language Models Infer Causation from Correlation?},
  author    = {Jin, Zhijing and Liu, Jiarui and Lyu, Zhiheng and Poff, Spencer and Sachan, Mrinmaya and Mihalcea, Rada and Diab, Mona and Sch{\"o}lkopf, Bernhard},
  booktitle = {Proceedings of the International Conference on Learning Representations (ICLR)},
  year      = {2024}
}

@article{moraffah2020causal,
author = {Moraffah, Raha and Karami, Mansooreh and Guo, Ruocheng and Raglin, Adrienne and Liu, Huan},
title = {Causal Interpretability for Machine Learning - Problems, Methods and Evaluation},
year = {2020},
issue_date = {June 2020},
publisher = {Association for Computing Machinery},
address = {New York, NY, USA},
volume = {22},
number = {1},
issn = {1931-0145},
url = {https://doi.org/10.1145/3400051.3400058},
doi = {10.1145/3400051.3400058},
journal = {SIGKDD Explor. Newsl.},
month = may,
pages = {18–33},
numpages = {16}
}

@ARTICLE{cheng2022evaluation,
  author={Cheng, Lu and Guo, Ruocheng and Moraffah, Raha and Sheth, Paras and Candan, K. Selçuk and Liu, Huan},
  journal={IEEE Transactions on Artificial Intelligence}, 
  title={Evaluation Methods and Measures for Causal Learning Algorithms}, 
  year={2022},
  volume={3},
  number={6},
  pages={924-943},
  doi={10.1109/TAI.2022.3150264}
  }

\appendix
\section{Methodology and Reproducibility Details}
\label{sec:methodology_details}

\subsection{LLM Evaluation Setup}
\label{sec:llm_evaluation}
Llama models were evaluated with temperature = 1.0 to maintain output diversity, while GPT models employed default temperature settings. For computational efficiency, we conducted preliminary screening on a 100-article subset to identify the most promising GPT variants before running the full-scale evaluation. Details about the GPT-4o-mini finetuning process and hyperparameters are provided in Section~\ref{sec:gpt_finetuning}.

\paragraph{Ideology Classification Prompt.} All LLMs (baseline GPT-4o-mini, Llama variants) were prompted with the following system instruction for ideology classification:

\begin{quote}
You are a political bias classifier specific to U.S.\ politics. Given the user's INPUT TEXT, return only valid JSON of the form \texttt{\{"bias":"left|center|right"\}}. No extra text.
\end{quote}

\noindent The article text was provided as the user message. For fine-tuned GPT-4o-mini, the same prompt structure was used during both training and inference.

\subsection{Topic Sentiment Extraction Details}
\label{sec:topic_sentiment_extraction}
Sentiment extraction uses two separate LLM calls per article:
\begin{enumerate}
    \item \textbf{Entity and Sentiment Analysis:} The \texttt{Llama-3.3-70b-versatile} model extracts named entities and key concepts from the article text and assigns each a continuous sentiment polarity score between $-1.0$ (strongly negative) and $+1.0$ (strongly positive). This model was selected for its sensitivity to contextual nuance relative to smaller alternatives.
    \item \textbf{Entity-to-Topic Association:} The \texttt{Llama3-70b-8192} model quantifies how strongly each extracted entity relates to predefined AllSides topic tags. The model handles ambiguity by distinguishing between a broad concept and specific incidents, for example, correctly separating terrorism as a general category from a specific event like the 2019 Christchurch shooting when assigning the entity ``shooting'' to a topic. Entity sentiment scores are then aggregated by topic tag to produce a per-topic sentiment profile for each article.
    \item \textbf{Top 5 Topic Selection:} The top 5 topics with the highest absolute sentiment scores are selected for each article. As the scores were aggregated at the previous step, this means that the final topic sentiment scores are no longer limited to having an absolute value $\leq 1$.
\end{enumerate}

To limit the risk of LLMs inferring bias from explicit sentiment signals, we used distinct sessions for the sentiment analysis and the ideology prediction steps.

\subsection{Community Detection and Experimental Settings}
\label{sec:community_detection}
The co-occurrence graph for Louvain community detection was constructed by filtering to edges with a minimum weight of five, meaning topic tag pairs that co-occurred in fewer than five articles were excluded. This threshold removes noise from outlier or one-off co-occurrences that may not reflect stable topical relationships. Self-loops allow single-tag articles to form their own communities. The algorithm yields a modularity score of 0.557 across 15 communities, with community sizes ranging from 1 to 90 tags.

\begin{table}[h]
    \centering
    \setlength{\tabcolsep}{2pt}\small\begin{tabular}{r|l}
    \toprule
    \textbf{Parameter} & \textbf{Setting} \\
    \midrule
    Estimator & LinearDML \\
    Sentiment Treatment Encoding & Continuous \\
    PCA Threshold & 0.95 \\
    Minimum Topic Presence & 50 \\
    Bootstrap Iterations & 2000 \\
    Bootstrap Random Seed & 42 \\
    Baseline GPT Model & \texttt{gpt-4o-mini-2024-07-18} \\
    \bottomrule
    \end{tabular}
    \caption{Experimental settings for causal analysis.}
    \label{tab:experimental_settings}
\end{table}

\subsection{GPT Finetuning}
\label{sec:gpt_finetuning}

\begin{table}[h!]
\centering
\begin{tabular}{r|l}
    \toprule
    \textbf{Hyperparameter} & \textbf{Values} \\
    \midrule
    Epochs & 3, 10 \\
    Learning Rate Multiplier & 1.8 \\
    Batch Size & 4 \\
\bottomrule
\end{tabular}
\caption{GPT-4o-mini fine-tuning hyperparameters.}
\label{tab:finetuning_hyperparameters}
\end{table}

\subsubsection{Experimental Variables and Configurations}

Table~\ref{tab:finetuning_hyperparameters} summarizes the hyperparameters used for fine-tuning the GPT-4o-mini model. We experimented with two different epoch counts (3 and 10) to evaluate the impact of training duration on model performance. We used the OpenAI recommended settings for learning rate multiplier (1.8) and batch size (4). We evaluate fine-tuning performance using 150, 300, 1,000, and 2,000 labeled examples, representing different points on the data efficiency curve. These sample sizes span the range from few-shot learning (150) to moderate supervision (2,000), enabling analysis of diminishing returns in training data scaling.

\subsubsection{Fine-tuning Protocol}
Input formatting follows the standard conversational template required by OpenAI models, with political bias labels converted to structured JSON responses to enable systematic evaluation. Each training example consists of the article text as user input and the corresponding political orientation (left/center/right) as the assistant response.

\begin{table}[ht]
\centering
\begin{tabular}{lc}
\toprule
\textbf{Configuration} & \textbf{F-1 Macro} \\
\midrule
Zero-shot Baseline & 51.86 \\
3 epochs, 150 samples & 33.37 \\
3 epochs, 300 samples & 25.04 \\
3 epochs, 1,000 samples & 69.23 \\
3 epochs, 2,000 samples & \textbf{69.37} \\
10 epochs, 150 samples & 28.03 \\
\bottomrule
\end{tabular}
\caption{GPT-4o-mini fine-tuning performance across different training configurations.}
\label{tab:llm_detailed_results}
\end{table}

We used the highest performing configuration (3 epochs, 2,000 samples) for the main causal analysis experiments. Table~\ref{tab:llm_detailed_results} provides a detailed breakdown of performance across all configurations, demonstrating the impact of training set size and input length on classification metrics.

\section{Community-Level ATE Breakdown}
\label{sec:community_ate}



\label{subsec:community_ate_complete}

We aggregate topics into the nine Louvain communities that pass the minimum-presence threshold (Section~\ref{sec:methodology}). The community sentiment $F_C$ is the mean sentiment over member topics present in an article (zero when none appear); we estimate the marginal ATE of each community's sentiment with a single \texttt{LinearDML} model over the community-aggregated treatment matrix, mirroring the topic-level setup. Aggregation increases articles per treatment unit but dilutes any single partisan signal, so effects are smaller than at the topic level.

\begin{figure}[t]
    \centering
    \includegraphics[width=\linewidth]{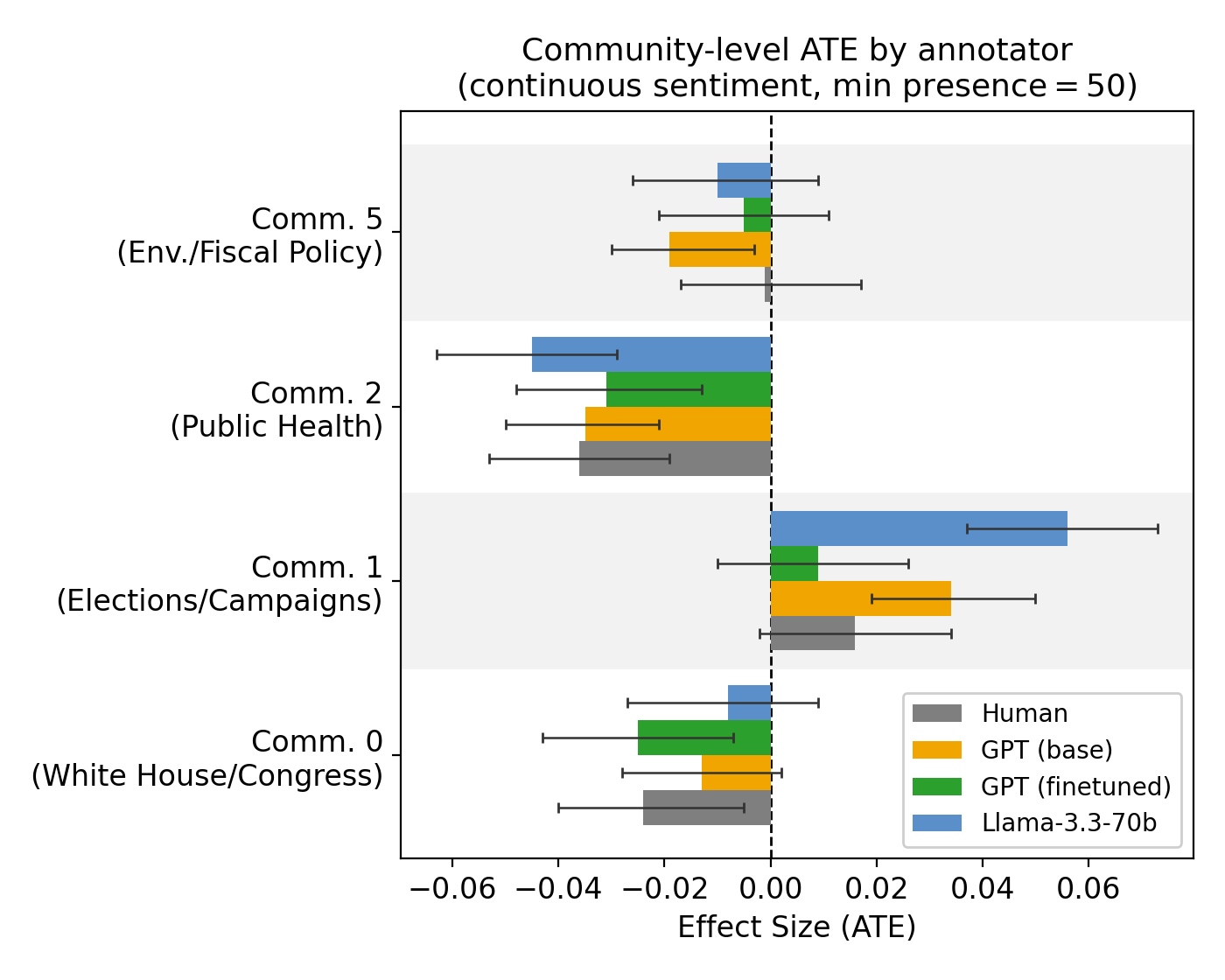}
    \caption{Community-level ATE by paradigm for the four communities with at least one significant effect. Error bars are 95\% bootstrap CIs. Full nine-community breakdown in Table~\ref{tab:community_ate}.}
    \label{fig:community_ate}
\end{figure}

Table~\ref{tab:community_ate} presents the ATE of community-level sentiment on ideology classification across all nine communities that met the minimum presence threshold (min presence${}=50$), for the four annotation paradigms.

\begin{table*}[t]
  \centering
  \small
  \setlength{\tabcolsep}{5pt}
  \resizebox{\textwidth}{!}{%
  \begin{tabular}{l c c c c c}
  \toprule
  \textbf{Community} & \textbf{Human} & \textbf{GPT (base)} & \textbf{Fine-tuned GPT} & \textbf{Llama-3.3-70B} & \textbf{N}\\
  \midrule
  Comm.\ 0 (White House/Congress) & $\mathbf{-.024}\,[-.040,-.005]$ & $-.013\,[-.028,+.002]$ & $\mathbf{-.025}\,[-.043,-.007]$ & $-.008\,[-.027,+.009]$ & 2473\\
  Comm.\ 1 (Elections/Campaigns)  & $+.016\,[-.002,+.034]$ & $\mathbf{+.034}\,[+.019,+.050]$ & $+.009\,[-.010,+.026]$ & $\mathbf{+.056}\,[+.037,+.073]$ & 3628\\
  Comm.\ 2 (Public Health)        & $\mathbf{-.036}\,[-.053,-.019]$ & $\mathbf{-.035}\,[-.050,-.021]$ & $\mathbf{-.031}\,[-.048,-.013]$ & $\mathbf{-.045}\,[-.063,-.029]$ & 1116\\
  Comm.\ 3 (Media/Technology)     & $+.003\,[-.017,+.020]$ & $-.003\,[-.019,+.014]$ & $-.000\,[-.017,+.018]$ & $+.006\,[-.014,+.025]$ & 890\\
  Comm.\ 4 (Culture/Education)    & $+.013\,[-.014,+.025]$ & $+.007\,[-.011,+.021]$ & $+.006\,[-.018,+.021]$ & $+.012\,[-.013,+.025]$ & 720\\
  Comm.\ 5 (Env./Fiscal Policy)   & $-.001\,[-.017,+.017]$ & $\mathbf{-.019}\,[-.030,-.003]$ & $-.005\,[-.021,+.011]$ & $-.010\,[-.026,+.009]$ & 1480\\
  Comm.\ 6 (Constitutional Rights)& $-.007\,[-.026,+.008]$ & $+.011\,[-.009,+.026]$ & $-.000\,[-.020,+.014]$ & $+.011\,[-.011,+.028]$ & 1094\\
  Comm.\ 8 (Foreign Policy)       & $+.013\,[-.004,+.029]$ & $+.000\,[-.011,+.013]$ & $+.009\,[-.008,+.024]$ & $+.011\,[-.006,+.026]$ & 2679\\
  Comm.\ 9 (Criminal Justice)     & $+.001\,[-.017,+.017]$ & $+.009\,[-.007,+.024]$ & $+.009\,[-.011,+.024]$ & $+.005\,[-.014,+.021]$ & 1364\\
  \bottomrule
  \end{tabular}}
  \caption{Community-level ATE by annotation paradigm (continuous sentiment, min presence${}=50$, min weight${}=2$). \textbf{Bolded} entries have 95\% CIs that exclude zero.}
  \label{tab:community_ate}
\end{table*}

Four of the nine communities produce at least one significant ATE. Community~2 (Public Health) is the standout: all four paradigms produce significant negative ATEs, making it the community-level analogue of the fully replicated partisan topics (e.g.\ Healthcare, Democratic Party) in Table~\ref{tab:multi-continuous-full}. Community~0 (White House and Congress) is significant under Human and fine-tuned GPT; Community~1 (Elections and Campaigns) is significant only under the two zero-shot paradigms; and Community~5 (Environment and Fiscal Policy) produces a single significant cell under GPT baseline. Directional agreement across paradigms is high: all four paradigms agree on the sign of the effect in each of the four significant communities. The remaining communities show small, non-significant effects consistent with the absence of a strong partisan signal among their constituent topics.

\section{Full Topic-Level ATE Results}
\label{sec:topic_level_results}

Table~\ref{tab:multi-continuous-full} reports the complete multi-treatment DML results for all 20 qualifying topics (summarised for the nine significant topics in Figure~\ref{fig:multi-continuous}), including the 95\% bootstrap CI ($B = 2{,}000$) for every ATE estimate under the four annotation paradigms. Topics are ordered by article count.

\begin{table*}[t]
  \centering
  \scriptsize
  \setlength{\tabcolsep}{4pt}
  \begin{tabular}{l c cccc}
    \toprule
    \textbf{Topic} & \textbf{N}
      & \textbf{Human}
      & \textbf{GPT (base)}
      & \textbf{Fine-tuned GPT}
      & \textbf{Llama-3.3-70B} \\
    \midrule
    Elections              & 1709 & $-.007\,[-.023,+.016]$ & $-.005\,[-.021,+.010]$ & $-.013\,[-.032,+.007]$ & $+.003\,[-.018,+.023]$ \\
    Politics               & 1520 & $-.006\,[-.022,+.012]$ & $+.007\,[-.008,+.023]$ & $-.012\,[-.029,+.005]$ & $+.006\,[-.011,+.025]$ \\
    Donald Trump           & 1424 & $\mathbf{+.043\,[+.025,+.059]}$ & $\mathbf{+.086\,[+.072,+.097]}$ & $\mathbf{+.045\,[+.027,+.063]}$ & $\mathbf{+.106\,[+.088,+.119]}$ \\
    Presidential Elections & 1331 & $+.004\,[-.017,+.022]$ & $+.009\,[-.005,+.024]$ & $+.003\,[-.017,+.022]$ & $+.008\,[-.011,+.029]$ \\
    White House            &  701 & $-.014\,[-.030,+.004]$ & $+.013\,[-.001,+.027]$ & $-.004\,[-.020,+.012]$ & $\mathbf{+.019\,[+.001,+.035]}$ \\
    World                  &  646 & $-.007\,[-.022,+.011]$ & $-.004\,[-.016,+.009]$ & $+.002\,[-.016,+.021]$ & $-.008\,[-.022,+.008]$ \\
    Immigration            &  485 & $-.010\,[-.024,+.008]$ & $+.011\,[-.004,+.031]$ & $-.009\,[-.022,+.007]$ & $+.017\,[-.000,+.040]$ \\
    Economy and Jobs       &  462 & $-.000\,[-.024,+.017]$ & $-.006\,[-.016,+.006]$ & $-.000\,[-.025,+.016]$ & $+.001\,[-.013,+.018]$ \\
    Coronavirus            &  441 & $-.004\,[-.021,+.011]$ & $-.007\,[-.018,+.004]$ & $-.004\,[-.024,+.010]$ & $-.012\,[-.026,+.001]$ \\
    Healthcare             &  373 & $\mathbf{-.029\,[-.045,-.011]}$ & $\mathbf{-.032\,[-.043,-.018]}$ & $\mathbf{-.031\,[-.046,-.011]}$ & $\mathbf{-.042\,[-.056,-.024]}$ \\
    Defense and Security   &  352 & $-.001\,[-.019,+.015]$ & $+.015\,[-.000,+.030]$ & $-.008\,[-.024,+.012]$ & $\mathbf{+.025\,[+.008,+.043]}$ \\
    Middle East            &  325 & $+.005\,[-.007,+.026]$ & $+.002\,[-.010,+.013]$ & $+.009\,[-.001,+.031]$ & $+.013\,[-.006,+.026]$ \\
    Public Health          &  297 & $-.010\,[-.024,+.006]$ & $+.002\,[-.008,+.014]$ & $-.009\,[-.026,+.008]$ & $+.002\,[-.011,+.015]$ \\
    Republican Party       &  293 & $\mathbf{+.037\,[+.023,+.052]}$ & $\mathbf{+.046\,[+.034,+.059]}$ & $\mathbf{+.028\,[+.013,+.042]}$ & $\mathbf{+.050\,[+.034,+.069]}$ \\
    Joe Biden              &  288 & $-.014\,[-.032,+.000]$ & $\mathbf{-.024\,[-.038,-.013]}$ & $\mathbf{-.022\,[-.038,-.004]}$ & $\mathbf{-.029\,[-.047,-.015]}$ \\
    Barack Obama           &  286 & $\mathbf{-.038\,[-.055,-.023]}$ & $\mathbf{-.049\,[-.060,-.035]}$ & $\mathbf{-.038\,[-.053,-.021]}$ & $\mathbf{-.057\,[-.069,-.040]}$ \\
    Hillary Clinton        &  279 & $\mathbf{-.044\,[-.063,-.027]}$ & $\mathbf{-.060\,[-.077,-.044]}$ & $\mathbf{-.048\,[-.067,-.031]}$ & $\mathbf{-.063\,[-.081,-.044]}$ \\
    Russia                 &  270 & $-.001\,[-.018,+.015]$ & $-.004\,[-.013,+.004]$ & $+.004\,[-.015,+.017]$ & $+.014\,[-.004,+.030]$ \\
    Democratic Party       &  261 & $\mathbf{-.046\,[-.059,-.031]}$ & $\mathbf{-.054\,[-.067,-.040]}$ & $\mathbf{-.047\,[-.059,-.033]}$ & $\mathbf{-.051\,[-.064,-.036]}$ \\
    Violence in America    &  260 & $-.007\,[-.025,+.011]$ & $-.000\,[-.013,+.016]$ & $-.002\,[-.019,+.018]$ & $-.005\,[-.020,+.014]$ \\
    \bottomrule
  \end{tabular}
  \caption{Full multi-treatment DML results: ATE per topic under continuous encoding (one-SD increase in sentiment score) with 95\% bootstrap CIs ($B = 2{,}000$). Topics ordered by article count. \textbf{Bolded} entries have CIs that exclude zero.}
  \label{tab:multi-continuous-full}
\end{table*}

Nine of the 20 topics produce at least one significant effect. Six are significant under all four paradigms (Donald Trump, Healthcare, Republican Party, Barack Obama, Hillary Clinton, Democratic Party) and Joe Biden under three; White House and Defense and Security reach significance only under Llama-3.3-70B. Effect significance tracks how polarising each annotator perceives the topic to be, rather than the topic's article count: the three largest topics by article count (Elections, Politics, Presidential Elections) produce no significant effects under any paradigm, while smaller but explicitly partisan topics are significant across paradigms. The significant partisan topics show uniform sign agreement, indicating that the annotators share a prior about these topics. Non-significant topics, by contrast, exhibit more directional disagreement across paradigms (e.g.\ Russia, where Human and GPT baseline estimate negative effects while fine-tuned GPT and Llama estimate positive ones), reflecting divergent annotator priors rather than corpus noise alone.

\section{Community- and Topic-Level Mediation Results}
\label{sec:community_mediation_results}


\subsection{Pathway Selection}

The pathways reported here are treatment--mediator pairs drawn from the three communities $\{0, 1, 2\}$ (White House and Congress, Elections and Campaigns, and Public Health respectively), for which all or nearly all paradigms produced significant ATEs in Appendix~\ref{sec:community_ate}. Of the six ordered pairs among these three communities, four satisfied the minimum-presence requirement (at least 50 articles tagged with both the treatment and mediator community) and produced non-degenerate mediation decompositions: the two configurations with Community~1 as treatment ($1 \to 0$, $1 \to 2$) and the two with Community~1 as mediator ($0 \to 1$, $2 \to 1$).

Confounders consist of all community sentiment scores except those of the treatment and mediator communities, reduced by (1) a presence filter dropping communities non-zero in fewer than 50 articles in the treatment subsample and (2) PCA retaining 95\% of variance. Because this depends only on the treatment subsample and not on ideology labels, it is identical across paradigms for a given treatment community.

\subsection{Full Community Mediation Results}

Table~\ref{tab:mediation_full_communities} reports TE, NDE, and NIE under the $Q_{25} \to Q_{75}$ contrast across the four paradigms and all four viable pathways. The $P_{10} \to P_{90}$ contrast yields directionally concordant results.

\begin{table*}[t]
\centering
\footnotesize
\setlength{\tabcolsep}{5pt}
\begin{tabular}{lcccccc}
\toprule
\textbf{Annotator} & \textbf{TE} & \textbf{95\% CI (TE)} & \textbf{NDE} & \textbf{95\% CI (NDE)} & \textbf{NIE} & \textbf{95\% CI (NIE)} \\
\midrule
\multicolumn{7}{l}{\textit{Community $1 \to 0$ ($N=3{,}628$, sentiment contrast $-0.3$ to $+0.2$)}} \\
\midrule
Human           & $+.018$ & $[-.044,+.080]$ & $+.017$ & $[-.008,+.043]$ & $+.001$ & $[-.005,+.010]$ \\
GPT (base)      & $+.039$ & $[-.013,+.091]$ & $\mathbf{+.039^*}$ & $[+.018,+.060]$ & $-.000$ & $[-.003,+.009]$ \\
Fine-tuned GPT  & $+.010$ & $[-.053,+.073]$ & $+.003$ & $[-.022,+.028]$ & $+.007$ & $[-.000,+.014]$ \\
Llama-3.3-70B   & $\mathbf{+.065^*}$ & $[+.001,+.129]$ & $\mathbf{+.069^*}$ & $[+.041,+.093]$ & $-.003$ & $[-.006,+.009]$ \\
\midrule
\multicolumn{7}{l}{\textit{Community $1 \to 2$ ($N=3{,}628$, sentiment contrast $-0.3$ to $+0.2$)}} \\
\midrule
Human           & $+.018$ & $[-.042,+.078]$ & $+.014$ & $[-.010,+.042]$ & $+.005$ & $[-.001,+.009]$ \\
GPT (base)      & $+.039$ & $[-.011,+.089]$ & $\mathbf{+.037^*}$ & $[+.018,+.060]$ & $+.002$ & $[-.001,+.007]$ \\
Fine-tuned GPT  & $+.005$ & $[-.056,+.065]$ & $-.000$ & $[-.022,+.027]$ & $+.005$ & $[-.002,+.007]$ \\
Llama-3.3-70B   & $\mathbf{+.067^*}$ & $[+.005,+.129]$ & $\mathbf{+.067^*}$ & $[+.041,+.092]$ & $-.000$ & $[-.001,+.009]$ \\
\midrule
\multicolumn{7}{l}{\textit{Community $0 \to 1$ ($N=2{,}743$, sentiment contrast $-0.2$ to $+0.1$)}} \\
\midrule
Human           & $-.021$ & $[-.083,+.041]$ & $\mathbf{-.030^*}$ & $[-.046,-.002]$ & $+.009$ & $[-.012,+.010]$ \\
GPT (base)      & $-.013$ & $[-.066,+.041]$ & $-.012$ & $[-.028,+.009]$ & $\mathbf{-.001^*}$ & $[-.021,-.003]$ \\
Fine-tuned GPT  & $-.026$ & $[-.089,+.036]$ & $\mathbf{-.034^*}$ & $[-.052,-.006]$ & $+.008$ & $[-.012,+.010]$ \\
Llama-3.3-70B   & $-.014$ & $[-.077,+.050]$ & $-.003$ & $[-.022,+.024]$ & $\mathbf{-.011^*}$ & $[-.031,-.008]$ \\
\midrule
\multicolumn{7}{l}{\textit{Community $2 \to 1$ ($N=1{,}116$, sentiment contrast $-0.3$ to $+0.1$)}} \\
\midrule
Human           & $-.081$ & $[-.182,+.020]$ & $\mathbf{-.081^*}$ & $[-.122,-.039]$ & $+.000$ & $[-.011,+.011]$ \\
GPT (base)      & $-.074$ & $[-.160,+.012]$ & $\mathbf{-.085^*}$ & $[-.120,-.055]$ & $+.011$ & $[-.009,+.015]$ \\
Fine-tuned GPT  & $-.071$ & $[-.179,+.037]$ & $\mathbf{-.059^*}$ & $[-.114,-.026]$ & $-.012$ & $[-.011,+.011]$ \\
Llama-3.3-70B   & $-.087$ & $[-.182,+.008]$ & $\mathbf{-.104^*}$ & $[-.141,-.065]$ & $+.017$ & $[-.014,+.013]$ \\
\bottomrule
\end{tabular}
\caption{Community-level mediation results ($Q_{25} \to Q_{75}$ contrast) across annotation paradigms. $^*$Significant at 95\% (CI excludes zero); such entries are \textbf{bolded}.}
\label{tab:mediation_full_communities}
\end{table*}

\begin{figure}[t]
    \centering
    \includegraphics[width=\linewidth]{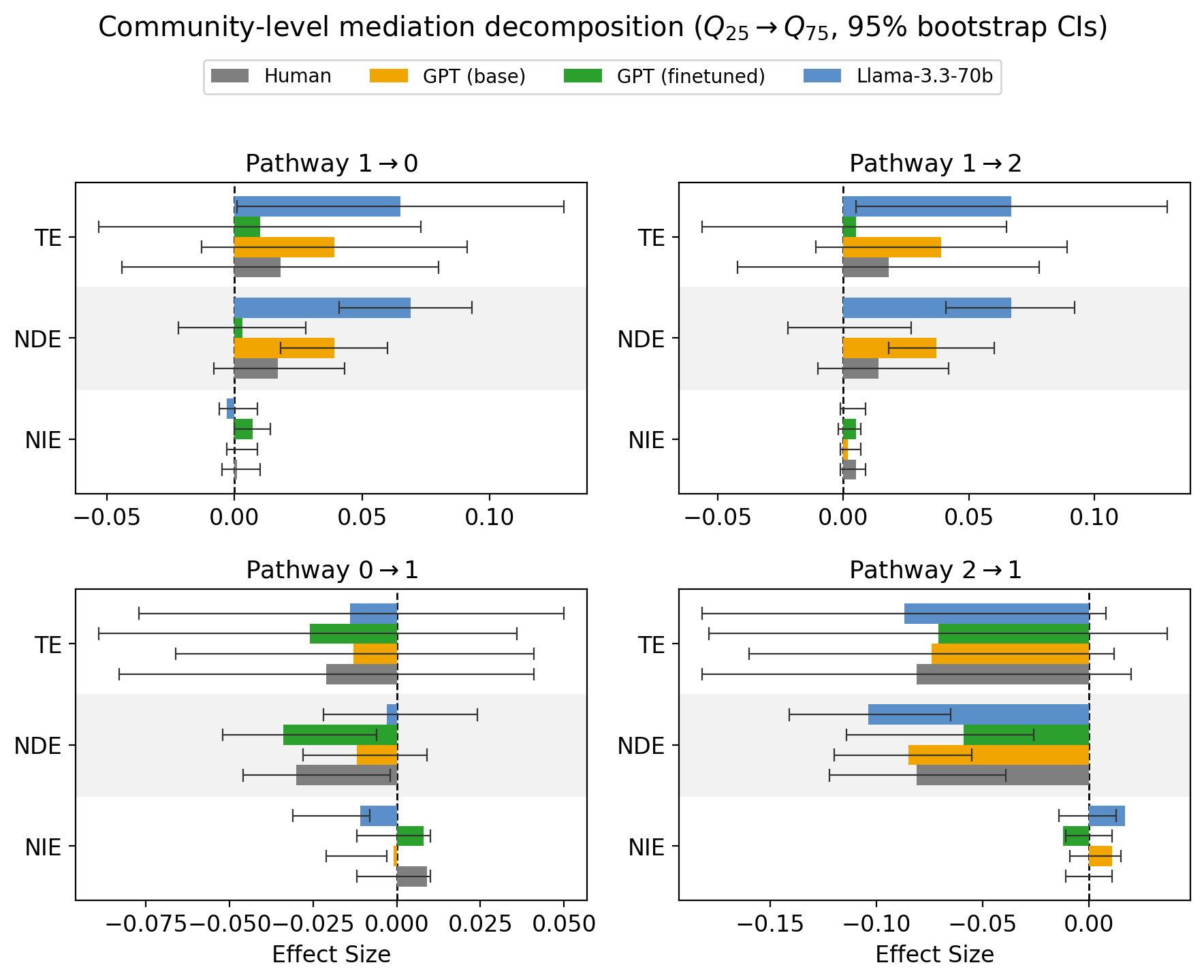}
    \caption{Community-level mediation decomposition ($Q_{25}\to Q_{75}$) for the four viable pathways among Communities 0 (White House/Congress), 1 (Elections/Campaigns), and 2 (Public Health). Error bars are 95\% bootstrap CIs for TE, NDE, and NIE.}
    \label{fig:community_mediation}
\end{figure}

Across all four pathways, significance concentrates in the Natural Direct Effect rather than the Natural Indirect Effect: sentiment toward one community shifts ideology predictions through a channel that does not flow through another community's sentiment. The $2 \to 1$ pathway (Public Health treatment) is the clearest case, with all four paradigms producing significant negative NDEs and negligible NIEs. The wide TE confidence intervals reflect both modest per-pathway sample sizes and the heterogeneity of pooling many topics of opposing partisan valence into a single community.

\subsection{Topic Mediation: Donald Trump Treatment}

Table~\ref{tab:te-trump} reports the Total Effect of Donald Trump sentiment on the ideology label for two percentile contrasts, decomposed into NDE and NIE.

\begin{table}[t]
  \centering
  \small
  \setlength{\tabcolsep}{4pt}
  \begin{tabular}{llrrr}
    \toprule
    \textbf{Paradigm} & \textbf{Contrast} & \textbf{TE} & \textbf{NDE} & \textbf{NIE} \\
    \midrule
    Human            & $Q25\!\to\!Q75$ & $+.135$ & $+.124$ & $+.010$ \\
                     & $P10\!\to\!P90$ & $+.314$ & $+.300$ & $+.014$ \\
    \addlinespace
    GPT (base)       & $Q25\!\to\!Q75$ & $\mathbf{+.255^*}$ & $+.249$ & $+.006$ \\
                     & $P10\!\to\!P90$ & $\mathbf{+.595^*}$ & $+.585$ & $+.010$ \\
    \addlinespace
    Fine-tuned GPT   & $Q25\!\to\!Q75$ & $\mathbf{+.158^*}$ & $+.141$ & $+.017$ \\
                     & $P10\!\to\!P90$ & $\mathbf{+.368^*}$ & $+.335$ & $+.032$ \\
    \addlinespace
    Llama-3.3-70B    & $Q25\!\to\!Q75$ & $\mathbf{+.316^*}$ & $+.316$ & $-.000$ \\
                     & $P10\!\to\!P90$ & $\mathbf{+.737^*}$ & $+.741$ & $-.004$ \\
    \bottomrule
  \end{tabular}
  \caption{Total Effect (TE) of Donald Trump sentiment on the ideology label for two percentile contrasts ($N = 1{,}424$). All four paradigms agree on a positive direction; three of four reach 95\% significance ($^*$, \textbf{bolded}), with Human narrowly missing. The direct channel (NDE) dominates the indirect channel (NIE) in all paradigms.}
  \label{tab:te-trump}
\end{table}

All four paradigms agree on a positive direction, and three of four reach significance under both contrasts; only Human narrowly misses. The direct channel dominates in every paradigm. The zero-shot LLMs (GPT baseline, Llama) produce the largest Total Effects, while fine-tuned GPT sits closest to the human estimate, indicating that fine-tuning attenuates rather than amplifies the Trump-sentiment signal.

\end{document}